\documentstyle[times]{article}

\newtheorem{example}{}

\newcommand{\startx} 
 {
  \begin{example} \rm \ \begin{minipage}[t]{12cm}}

\newcommand{\startxl}[1] 
  {
  \begin{example} \label{#1} \ \rm \begin{minipage}[t]{11cm}}

\newcommand{\startxll}[1] 
  {
  \begin{example} \label{#1} \ \rm \begin{minipage}[t]{12cm}}

\newcommand{\startpx} 
 {
  \begin{example} \rm \ \begin{minipage}[t]{15cm}}

\newcommand{\startpxl}[1] 
 {
  \begin{example} \label{#1} \ \rm \begin{minipage}[t]{11cm}}

\newcommand{\stopx} 
 {\end{minipage} \end{example} 
	\vspace{\topsep}
}

\newcommand{\startfxl}[1] 
 {\begin{figure}[p]
  \vspace{-.5in}
  \begin{example}
  \label{#1} \rm \
  \begin{minipage}[t]{15cm}}

\newcommand{\startfxll}[1] 
 {\begin{figure}
  \begin{example}
  \label{#1} \rm \
  \begin{minipage}[t]{11cm}}

\newcommand{\startfxlt}[1] 
 {\begin{figure}[t]
  \begin{example}
  \label{#1} \rm \
  \begin{minipage}[t]{12cm}}

\newcommand{\stopfx} 
 {\stopx \end{figure}}

\newenvironment{lexlist}{\begin{list}{}
{\setlength{\itemsep}{0ex}\setlength{\parsep}{\parskip}
\setlength{\leftmargin}{1.5em}\setlength{\labelsep}{.5em}
\setlength{\labelwidth}{1em}}}{\end{list}}

\def\thebibliography#1{\subsection{References\markboth  
  {References}{References}}\vspace{-.3cm}\list
  {[\arabic{enumi}]}{\settowidth\labelwidth{[#1]}\leftmargin\labelwidth
    \advance\leftmargin\labelsep\usecounter{enumi}}}

\renewcommand{\section}[1]
{\nopagebreak
 \par
 \noindent\parbox{\hsize}
 {\setcounter{subsection}{0}
  \addtocounter{section}{1}
  \begin{center}
   {\bf \thesection . #1 }
  \end{center}}
\noindent
}

\newcommand{\baresection}[1]
{\nopagebreak
 \par
 \noindent\parbox{\hsize}
 {\vspace{1ex}
  \begin{center}
   {\bf #1}
  \end{center}}
\noindent
}

\newcommand{\tb}[1]{\begin{tabbing} #1 \end{tabbing}}

\makeatletter
\newcommand{\citedelimiters}[2]{    
\def\@cite##1##2{#1{##1\if@tempswa , ##2\fi}#2}
\def\@biblabel##1{#1##1#2\hfill}
}
\makeatother

\citedelimiters{}{}

\renewcommand{\bibitem}[2]{
\write1{\string\bibcite{#2}{\mbox{#1}}}
\item[]}

\title{{\Large Specifying Intonation from Context for Speech Synthesis}}

\author{{\large Scott Prevost and Mark Steedman}\\
{\footnotesize\em Computer and Information Science, University of
Pennsylvania, Philadelphia, PA 19104-6389, USA}\\
\ \\
{\normalsize To appear in {\em Speech Communication}.}\\
{\normalsize Accepted April 26, 1994.
Revised May 16, 1994.}}

\date{}

\begin{document}
\maketitle
\baresection{Abstract}
\small
This paper presents a theory and a computational implementation for
generating prosodically appropriate synthetic speech in response to
database queries.  Proper distinctions of contrast and emphasis are
expressed in an intonation contour that is synthesized by rule under
the control of a grammar, a discourse model, and a knowledge base.
The theory is based on Combinatory Categorial Grammar, a formalism
which easily integrates the notions of syntactic constituency,
semantics, prosodic phrasing and information structure.  Results from
our current implementation demonstrate the system's ability to
generate a variety of intonational possibilities for a given sentence
depending on the discourse context.

\vspace{2.5ex}\noindent
Cet article vise a pr\'esenter une th\'eorie et une r\'ealisation informatique
de la g\'en\'eration de la par\^ole synth\`ethique accompagn\'ee d'intonation
appropri\'e, en r\'eponse \`a des enqu\^etes apropos d'une base de donn\'ees.
Les distinctions appropri\'ees de contraste et d'emphase sont marqu\'ees par
l'intonation automatiquement synthesis\'e sous la gouvernance de la grammaire,
un mod\`ele du discours, et d'une r\'epr\'esentation de la domaine cognitive.
La th\'eorie se fonde sur la Grammaire Categoriale Combinateurique, formalisme
qui se pr\^ete \`a l'int\'egration directe de la syntaxe, la semantique, la
structure prosodique, et le statut discursale de l'information.  Les
r\'esult\^ats de nos exp\'eriences demontrent les capacit\'es du syst\`eme de
g\'en\'erer plusieurs intonations diff\'eremment modul\'es selon le contexte du
discours pour une phrase donn\'ee.

\vspace{2.5ex}\noindent
Dieser Artikel pr\"{a}sentiert ein Modell zur Generierung prosodisch
ad\"{a}quater, synthetisierter Antworten auf Datenbankanfragen.  Dabei
werden die passenden Unterscheidungen zwischen Kontrast und Betonung
in Bezug auf ein Diskursmodell und eine 
Wissensbasis vermittelt.  Das Modell f\"{u}r die Generierung der
Betonungen basiert auf Combinatory Categorial Grammar (Kombinatoriale
Kategorial-Grammatiken), ein Formalismus, der die Verwendung von
syntaktischen Konstituenten, prosodischer Phrasierung und
Informationsstrukturen integriert.  Resultate unserer Implementierung
demonstrieren die F\"{a}higkeit des Systems, eine breite Auswahl von
Intonationsm\"{o}glichkeiten f\"{u}r einen gegebenen Satz in Abh\"{a}ngigkeit
vom Diskurs-Kontext zu generieren.
\newpage

\normalsize

\section{Introduction}
One source of unnaturalness in the output of many text-to-speech systems
stems from the involvement of algorithmically generated default
intonation contours, applied under minimal control from syntax and
semantics.  The intelligibility of the speech produced by these
systems is a tribute to both the resilience of human language
understanding and the ingenuity of the algorithms' inventors.  It
has often been noted, however, that the results frequently sound
unnatural when taken in context, and may on occasion mislead the
hearer.

It is for this reason that a number of discourse-model-based speech
generation systems have been proposed, in which intonation contour is
determined from context or the model.  Work in this area includes an
early study by Young and Fallside (\cite{Youn:79}), and studies by
Terken (\cite{Terk:84}), Houghton (\cite{Houg:86}), Isard and
Pearson (\cite{Isar:88}), Davis and
Hirschberg (\cite{Davi:88}), Hirschberg (\cite{Hirs:90}), and
Zacharski {\em et al.}
(\cite{Zach:93}), although the representations of information structure
and its relation to syntax employed by these authors are rather different
{}from those proposed here.

Consider the exchange shown in \ref{ex:crux}, which is an artificial
example modeled on the domain of TraumAID, a medical expert system in
the context of which we are investigating spoken language
output.\footnote{The examples used throughout the paper are based on a
the domain of TraumAID, which is currently under development at the
University of Pennsylvania (Webber {\em et al.} \cite{webb:92}).  The
lay reader may find it useful to know that a {\em thoracostomy} is the
insertion of a tube into the chest, and {\em pneumothorax} refers to
the presence of air or gas in the pleural cavity.} This particular
example is slightly unrealistic in that TraumAID acts purely as a
critiquing device and does not possess such an interactive query
system for its knowledge base; nor is it likely that such a query
system would be of practical use in the trauma surgery.  However,
such examples are useful for present purposes since they force
unambiguously contrastive contexts that motivate intonational focus
and contrastive stress.

In example \ref{ex:crux}, capitals indicate stress and brackets
informally indicate the intonational phrasing.  The intonation contour
is indicated more formally using a version of Pierrehumbert's notation
(cf. Pierrehumbert \cite{Pier:80}, Pierrehumbert and Hirschberg
\cite{BandP:86}).\footnote{A brief summary of Pierrehumbert's notation
can be found in Steedman (\cite{sandi}).} In this notation, L+H* and
H* are different high pitch accents. LH\% (and its relative LH\$) and
L (and its relatives LL\% and LL\$) are rising and low boundaries
respectively.  The difference between members of sets like L, LL\% and
LL\$ boundaries embodies Pierrehumbert and Beckman's (\cite{BandP:86})
distinction between intermediate phrase boundaries, intonational
phrase boundaries, and utterance boundaries.\footnote {Since utterance
boundaries always coincide with an intonational phrase boundary, this
distinction is often left implicit in the literature, both being
written with \% boundaries.  For purposes of synthesis, however, the
distinction is important.} We shall skate over the former distinction
here, noting only that utterance boundaries are distinguished from the
others by a greater degree of lengthening and pausing.

The other annotations in \ref{ex:crux} indicate that the intonational
tunes L+H*~LH\% (or the related L+H*~LH\$) and H*~L (or the related
H*~LL\$) convey two distinct kinds of
discourse information.  First, both H* and L+H* pitch accents mark the
word that they occur on (or rather, some element of its
interpretation) for ``focus'', which in
the context of such simple queries as example \ref{ex:crux} usually
implies contrast of some kind.  Second, the tunes as a whole mark the
constituent that bears them (or rather, its interpretation) as having
a particular function in the discourse.  We have argued at length
elsewhere that, at least in this same restricted class of dialogues,
the function of the L+H*~LH\% and L+H*~LH\$ tunes is to mark the
``theme'' -- that
is, ``what the participants have agreed to talk about''.  The
\mbox{H*~L(L\%/\$)} tune marks the ``rheme'' -- that is, ``what the speaker has
to say'' about the theme.  This phenomenon is a strong one: the same
intonation contour sounds quite anomalous in the context of a question
that does not establish an appropriate theme, such as ``which
procedure is needed for the persistent {\sc pneumothorax}?''.  The
advantage for present purposes of Pierrehumbert's system, like other
autosegmental approaches, is that the entire tune
can be defined independently of the particular string that it occurs
with, by interpolation of pitch contour between the pitch-accent(s)
and the boundary for those parts bearing no tonal annotation.  It
will be notationally convenient to speak of the latter as bearing
``null tone''.  (Of course such elements may
bear pitch and even secondary accent, and the
specification of such details of the interpolated contour is by no
means a trivial matter.
However, we do not believe that anything hangs crucially on our use of
this theory of intonation, rather than some other.)

\startfxll{ex:crux}
\small
\begin{lexlist}
\item [Q:] I know that a {\sc left} thoracostomy is needed for the {\sc simple}
pneumothorax,\\
\setlength{\tabcolsep}{.15em}
\begin{tabular}[t]{ccccccc}
(But & what & condition)& (is a &{\sc right}& thoracostomy & needed for?)\\
     & L+H* & LH\% & & H* & & LL\$
\end{tabular}
\item [A:]
\setlength{\tabcolsep}{.15em}
\begin{tabular}[t]{cccccccccccc}
   (A & & {\sc right} & & thoracostomy & is needed for) & & (the & &
{\sc persistent } & & pneumothorax.)\\
         &  &  L+H* & &    & LH\%  &     &  &  & H*  &     & LL\$\\
\cline{1-1} \cline{3-3} \cline{5-6} \cline{8-8} \cline{10-10} \cline{12-12}
{\em ground} & & {\em focus} & & \multicolumn{2}{c}{\em ground} &
& {\em ground} & & {\em focus} & &{\em ground}\\
\cline{1-6} \cline{8-12}
\multicolumn{6}{c}{\em Theme}& &\multicolumn{5}{c}{\em Rheme}
\end{tabular}
\end{lexlist}
\stopfx

\section{Combinatory Prosody}
{}From the example in the preceding section, it is clear that
intonational units corresponding to theme or rheme
need not always correspond to a traditional syntactic
constituent.  Since many problems in the analysis and synthesis of spoken
language result from this apparent independence of syntactic and
intonational phrase boundaries, we have chosen to base our system
on Combinatory Categorial Grammar (CCG), a formalism that
generalizes the notion of surface constituency, allowing multiple
derivations and constituent structures for sentences, including ones in
which the subject and verb of a transitive sentence can exist as a
constituent, complete with an interpretation.

CCG (Steedman \cite{cgpg,gacc,acl90,sandi}) is an extension of
Categorial Grammar (CG).  Elements like verbs are associated with a
syntactic ``category'' which identifies them as {\em functions}, and
specifies the type and directionality of their arguments and the type
of their result.  We use a notation in which a rightward-combining
functor over a domain $\beta$ into a range $\alpha$ is written $\alpha
/ \beta$, while the corresponding leftward-combining functor is written
$\alpha\backslash\beta$.  $\alpha$ and $\beta$ may themselves be
function categories.   For example, a transitive verb is a function
{}from (object) NPs into predicates -- that is, into functions from
(subject) NPs into S, written as follows:
\startxl{ex:prefers1}
$(S\backslash NP)/NP$
\stopx
We also need the following two rules of functional application, where
$X$ and $Y$ are variables over categories:
\startxl{ex:Arules}
{\sc Functional Application:}\\
$\begin{array}[t]{lccccc}
a. &  X/Y &  Y &  \Rightarrow &  X  & (\verb/>/)\\
b. &  Y   &  X\backslash Y &  \Rightarrow &  X & (\verb/</)
\end{array}$
\stopx
These rules allow the function category \ref{ex:prefers1} to combine
with arguments to yield context-free derivations of which \ref{ex:basic1}
is a simple example:\footnote{It may be helpful for the
reader to know that {\em lavage} refers to the therapeutic cleansing
of an organ.}
\startxl{ex:basic1}
\small
\begin{verbatim}
Traumaid   recommends   lavage
--------   ----------   ------
   NP      (S\NP)/NP      NP
           ------------------->
                   S\NP
-----------------------<
           S
\end{verbatim}
\stopx

The syntactic types in this derivation are simply a reflection of the
corresponding semantic
types, apart from the addition of directional information.  If we
expand the category \ref{ex:prefers1} to express the semantics of the
transitive verb,
the same context-free derivation can be made to build a
compositional interpretation, $({\em recommend}'\ {\em lavage}')\ {\em
traumaid}'$.  One way of writing such an interpreted category that is
particularly convenient for translating into unification-based
programming languages like Prolog is the following:
\startxl{ex:prefers2}
$\it (S:recommend'~x~y\backslash NP:y)/NP:x$
\stopx
In \ref{ex:prefers2}, syntactic types are paired with a semantic
interpretation via the colon operator, and the category is that of a
function from NPs (with interpretation $\it x$) to functions
{}from NPs (with interpretation $\it y$) to Ss (with
interpretation $\it recommend'~x~y$).   Constants in
interpretations bear primes, variables do not, and there is a
convention of left-associativity, so that $recommend'~x~y$ is
equivalent to $(recommend'~x)~y$.

CCG extends this strictly context-free categorial base in two respects.  First,
all arguments, such as NPs, bear only {\em type-raised} categories, such as
$S/(S\backslash NP)$.  That is to say that the category of an NP,
rather than being that of a simple argument, is that of a function
over functions-over-such-arguments, namely verbs and the like.
Similarly, all functions into such categories, such as
determiners, are functions into the raised categories, such as
$(S / (S \backslash NP)) /N$.  For example, subject NPs bear the
following category in the full notation:
\startxl{ex:full}
traumaid := $\it S:s/(S:s\backslash NP:traumaid')$
\stopx
The derivation of the same simple transitive sentence using
type-raised categories is illustrated in example \ref{ex:basic2}
in the abbreviated notation.\footnote {It is important to
realize that the semantics of the type raised categories and of the
application rules ensures that this derivation yields an S with the
same interpretation as the earlier derivation \ref{ex:basic1}, namely
$\it recommend'~lavage'~traumaid'$.  At first glance, it looks as
though type-raising will expand the lexicon alarmingly.  One way round
this problem is discussed in
Steedman (\cite{trandd}).}
\startxl{ex:basic2}
\small
\begin{verbatim}
Traumaid  recommends        lavage
--------  ----------  ------------------
S/(S\NP)  (S\NP)/NP   (S\NP)\((S\NP)/NP)
          ------------------------------<
                     S\NP
------------------------->
           S
\end{verbatim}
\stopx

Second, the combinatory rules are extended to include {functional
composition}, as well as application:
\startxl{ex:>B}
{\sc Forward Composition} (\verb|>B|):\\
$\begin{array}{cccc}
X/Y & Y/Z & \Rightarrow _{{\bf B}} & X/Z
\end{array}$
\stopx
This rule allows a {\em
second} syntactic derivation for the above sentence, as
shown in example \ref{ex:basic3}.\footnote
{As before, it is important to realize that the semantics of the
categories and of the new rule of functional composition guarantee
that the S
yielded in this derivation bears exactly the same interpretation as
the original purely applicative derivation \ref{ex:basic1}.}
\startxl{ex:basic3}
\small
\begin{verbatim}
Traumaid  recommends   lavage
--------  ----------  --------
S/(S\NP)  (S\NP)/NP   S\(S/NP)
-------------------->B
        S/NP
        ----------------------<
                 S
\end{verbatim}
\stopx

The original reason for making these moves was to capture the fact
that fragments like {\em Traumaid recommends}, which in traditional
terms are not regarded as syntactic constituents, can nevertheless
take part in coordinate constructions, like \ref{ex:examples}a, and
form the residue of relative clause formation, as in
\ref{ex:examples}b.
\startxl{ex:examples}
\begin{lexlist}
\item [a.] You propose, and {\em Traumaid recommends}, lavage.
\item [b.] The treatment that {\em Traumaid recommends}
\end{lexlist}
\stopx
The full extent of this theory (which covers unbounded rightward and
leftward ``movement'', and a number of other types of supposedly
``non-constituent'' coordination), together with the general class of
rules from which the composition rule is drawn, and the problem of
processing in the face of such associative rules, is discussed in the
earlier papers, and need not concern us here.  The point for present
purposes is that the partition of the sentence into the object and a
non-standard constituent ($\it S:recommend'~x~traumaid'/NP:x$) makes
this theory structurally and semantically perfectly suited to the
demands of intonation, as exhibited in exchanges like the
following:\footnote {A similar argument in a related categorial
framework is made by Moortgat (\cite{Moor:89}).}
\startxl{ex:exchange}
\begin{lexlist}
\item [Q:] I know that the surgeon recommends a left thoracotomy,\\
but what does Traumaid recommend?
\item [A:] \tb{({\sc t}\={\sc raumaid} recomm\=ends) ({\sc la}\={\sc vage}.)\\
\> L+H* \> LH\% \>  H* LL\$}
\end{lexlist}
\stopx

We can therefore directly incorporate intonational constituency in
syntax, as follows (cf. Steedman \cite{acl90,sandi,brussels}).  First,
we assign to each constituent an autonomous prosodic category,
expressing its potential for combination with other prosodic
categories.  Then we lock these two structural systems together via
the following principle, which says that syntactic and prosodic
constituency must be isomorphic:
\startpxl{ex:pcc}
{\sc Prosodic Constituent Condition:}\\
Combination of two syntactic categories via a syntactic combinatory
rule is only allowed if their prosodic categories can also combine
via a prosodic combinatory rule.
\stopx

One way to accomplish this is to give pitch accents the category of
functions from boundaries to intonational/intermediate phrases.
As in CCG, categories consist of a (prosodic) structural type, and an
(information structural) interpretation, associated via a colon.  The pitch
accents have the following functional types:\footnote
{Here we are ignoring the possibility of multiple pitch accents in the
same prosodic phrase, but cf. Steedman (\cite{sandi}).}
\startxl{ex:pitchcats}
\begin{tabular}[t]{lll}
L+H* & := & ${\em p:theme/b:lh}$\\
H*   & := & ${\em p:rheme/b:ll}$
\end{tabular}
\stopx
We further assume, following Bird (\cite{Bird:91}),
that the presence of a pitch accent causes some element(s) in the
translation of the category to be marked as focused, a matter which
we will for simplicity assume to occur at the level of the lexicon.  For
example, when {\it recommends} bears a pitch accent, its category will
be written as follows:
\startxl{ex:recommends*}
$ (S:*recommend'~x~y\backslash NP:y)/NP:x$
\stopx

We depart from earlier versions of this theory in assuming that
boundaries are not simply {\em arguments} of such functions, but are
rather akin to {\em type-raised} arguments, as follows:\footnote {Note
again that \$ boundaries are often conflated with \% intonational
phrase boundaries in the literature.  These categories, which in some
sense imply that boundaries are phonological heads, constitute a
modification to previous versions of the present theory that brings it
more closely into line with the proposals in Pierrehumbert and
Hirschberg (\cite{PandH:90}).  The idea that boundaries are functors
has been independently proposed by Kirkeby-Garstad and Polgardi
(p.c.).}
\startxl{ex:phrasalcats2}
\begin{tabular}[t]{lll}
L   & := & $p:rheme\backslash (p:rheme/b:ll)$\\
LL\$ & := & $u:rheme\backslash (p:rheme/b:ll)$\\
LH\%   & := & $p:theme\backslash (p:theme/b:lh)$\\
LH\$ & := & $u:theme\backslash (p:theme/b:lh)$
\end{tabular}
\stopx
These categories closely correspond to Pierrehumbert's distinction
between  various levels of phonological phrases.  For example, the
boundary L maps an
H* pitch accent into an intermediate phrase rheme, $p:rheme$.
The LH\% boundary maps an L+H* pitch accent onto a full intonation
phrase, which it is convenient for present purposes to write as
$p:theme$.   (In a fuller notation we would make the distinction
between intermediate and intonational phrases explicit, but for
present purposes it is irrelevant).  The LH\$ boundary
maps the same L+H* pitch accent
into an utterance-level thematic phrase, written $u:theme$.

The categories that result from the combination of a pitch
accent and a boundary may or may not
constitute entire prosodic phrases, since there may be
prenuclear material bearing null tone.  There may also be
material bearing null tone separating the
pitch accent(s) from the boundary.  (Both possibilities are
illustrated in \ref{ex:crux}).   We therefore assign the following
category to the null tone, which can thereby
apply to the right to any non-functional category of the form
$X:Y$, and compose to the right with any function into such a
category, including another null tone, to yield the same category:
\startxl{ex:nullcat}
\begin{tabular}[t]{lll}
\O & := & ${X:Y}/{X:Y}$
\end{tabular}
\stopx
It is this omnivorous category that allows intonational tunes to be
spread over arbitrarily large constituents, since it allows the pitch
accent's desire for a boundary to propagate via composition into the
null tone category, as in the earlier papers.

In order to allow the derivation to proceed above the level of
complete prosodic phrases identifying themes and rhemes, we need the two
unary category-{\em changing} rules shown in \ref{ex:minor} and
\ref{ex:major} to change the phonological category of complete themes
and rhemes.\footnote {These rules represent another minor
departure from the earlier papers.}
\startxl{ex:minor}
\begin{tabular}[t]{ccc}
$\Sigma$ & $\Rightarrow$ & $\Sigma$\\
$p:X$ & & $utterance/utterance$
\end{tabular}
\stopx
\startxl{ex:major}
\begin{tabular}[t]{ccc}
$\Sigma$ & $\Rightarrow$ & $\Sigma$\\
$u:X$ & & $utterance$
\end{tabular}
\stopx
These rules change the prosodic category either to
$utterance$, or to an endocentric function over that category.
These types capture the fact that the
LL\$ and LH\$ boundaries can only occur at the end of a sentence, thereby
correcting an overgeneration in some early versions of this theory
noted by Bird (\cite{Bird:91}).  The fact that
$utterance$ is an atom rather than a
term of the form $X:Y$ is important, since it means that it can
unify only with another $utterance$.   This is vital to the preservation of the
intonation structure.\footnote
{The category has the effect of preventing further composition into the
null tone achieved in the earlier papers by a restriction on forward
prosodic composition.}

The application of the above two rules to a complete intonational
phrase should be thought of as precipitating a side-effect whereby a
copy of the category $\Sigma$ is associated with the clause as its
theme or rheme.  (We gloss over details of how
this is done, as well as a number of further
complications arising in sentences with more than one rheme).

In Steedman (\cite{sandi,brussels}), a related set of rules of which
the present ones form a subset are shown to be well-behaved with a
wide range of examples.  Example \ref{ex:fredate} gives the derivation
for an example related to \ref{ex:basic3}.\footnote{Note that since the raised
object category is not crucial, it has been replaced by NP for ease of reading
comprehension.  Also note the focus-marking effect of the pitch accents.}
\startfxll{ex:fredate}
\footnotesize
\begin{verbatim}
      Traumaid                  recommends             lavage
        L+H*                        LH%                  H* LL$
-----------------------  --------------------------  -----------
S:s/(S:s\NP:*traumaid') (S:recommend'x y\NP:y)/NP:x  NP:*lavage'
     p:theme/b:lh          p:theme\(p:theme/b:lh)      u:rheme
------------------------SYN----------------------->B
------------------------PHON----------------------<
           S:recommend' x *traumaid'/NP:x
                      p:theme
           =============PHON=============            ====PHON===
           S:recommend' x *traumaid'/NP:x            NP:*lavage'
                 utterance/utterance                  utterance
           -------------------------SYN------------------------->
           -------------------------PHON------------------------>
                      S: recommend' *lavage' *traumaid'
                                 utterance
\end{verbatim}
\begin{lexlist}
\item [Theme:] $S:recommend'~z~{*traumaid'}/NP:z$
\item [Rheme:] $NP:*lavage$
\end{lexlist}
\stopfx
Note that it is the identification of the theme and rheme at
the stage {\em before} the final reduction
that determines the information structure for the response, for it is
at this point that
discourse elements like the theme of the answer can be defined, and can be
used in semantically-driven synthesis of intonation contour directly
{}from the grammar.

Of course, such effusively informative intonation contours are
comparatively rare in normal dialogues.  A more usual response to the
question ``What does Traumaid recommend?'' in \ref{ex:exchange} would
put low pitch -- that is, the null tone in Pierrehumbert's terms -- on
everything except the focus of the rheme, {\em lavage}, as in
\ref{ex:lavage}.
\startxl{ex:lavage}
\tb{Traumaid recommends {\sc la}\={\sc vage}.\\
\> H* LL\$}
\stopx
Such an utterance is of course ambiguous as to whether the theme is
{\em traumaid} or {\em what traumaid recommends}.  The earlier papers show
that such ``unmarked'' themes, which include no primary pitch accent because
they are entirely background, can be captured by a ``Null Theme
Promotion Rule'', as follows:
\startxl{ex:nulltheme}
$\begin{array}[t]{ccc}
\Sigma &   & \Sigma \\
{X:Y}/{X:Y} & \Rightarrow & p:theme
\end{array}$
\stopx
This rule says that any sequence bearing the null tone can be
regarded as an ``unmarked'' intermediate phrase theme.

\section{Modeling Contrast}
The preceding remarks about the ambiguity of unmarked themes should
make it clear that in general the information structure of the
response to a query cannot be identified on the basis of the question
alone, but requires information from the discourse model as well, to
which we now turn.\footnote {See Prevost and Steedman (\cite{eacl:93})
for an investigation of how much one can get away with on the basis of
the question alone.}

This remark applies even more strongly to the assignment of focus and
the corresponding pitch accents in the generation of the response, as
Davis and Hirschberg (\cite{Davi:88}), and Hirschberg
(\cite{Hirs:90}), among others, have pointed out.  That is, while it
might appear as though pitch-accents could be assigned on some basis
such as the mention or non-mention of the relevant words in the theme
of the query, such an expedient will often break down.  Consider the
following example, which might be produced by such a strategem, since
the words ``left'' and ``thoracotomy'' do not occur in the theme {\em Which
incision}:\footnote{It may be helpful to point out that a {\em
thoracotomy} is a surgical incision of the chest wall, and a {\em
thoracostomy} is the insertion of a tube into the chest.}
\startxl{ex:error}
\begin{lexlist}
\item[Q:] Which incision does {\sc traumaid} prefer?
\item[A:] \tb{({\sc t}\={\sc raumaid} pre\=fers) (a {\sc l}\={\sc eft}
thora{\sc c}\={\sc ot}omy.)\\
\> L+H* \> LH\% \> H* \> H* LL\$}
\end{lexlist}
\stopx
In some contexts, including the null context, this intonation contour
will indeed be appropriate.  However, in any context where thoracotomy
procedures are already established as the set of procedures in question,
the pitch accent on {\em thoracotomy} in
the response will be inappropriate and perhaps even misleading.

For example, in \ref{ex:contrast} below, the noun {\em thoracotomy}
must remain unstressed while the adjective {\em left} must be accented in
the response, despite having been explicitly mentioned in the text of
the question.\footnote {In using these examples to motivate the
treatment of contrast in the system, we go beyond the class of
discourses that are actually handled by the system as currently
implemented.  We are in fact glossing over a number of subtle problems
concerning the theme-rheme structures that are involved, and the
precise reflection of these information structures in intonation.}
Here the question itself establishes a contextual set.  The fact that
the entity that is referenced in the response must be contrasted with
other alternatives in this set on the relevant property requires the
assignment of a pitch accent to the corresponding word.
\startxl{ex:contrast}
\begin{lexlist}
\item[Q:] Does Traumaid prefer a {\sc left} thoracotomy or a {\sc
right} thoracotomy?
\item[A:] (Traumaid prefers) (a {\sc left} thoracotomy.)
\end{lexlist}
\stopx
The mere fact that alternatives are contrasted on a given
property is not enough however to mandate the inclusion of a pitch accent on
the corresponding linguistic material.   The property in question
must restrict contrastively {\em at the relevant point in the semantic
evaluation}, before a pitch accent is forced.
Thus, in a situation in which the choices include a left thoracotomy,
a right thoracotomy, a left thoracostomy and a right thoracostomy,
the response to question \ref{ex:contrast2},
in which the adjective is unstressed, is perfectly
appropriate:\footnote
{That is not to claim that the adjective {\em cannot} carry a pitch
accent, of course.}
\startxl{ex:contrast2}
\begin{lexlist}
\item[Q:] Does Traumaid prefer a {\sc left} thora{\sc cot}omy or a
{\sc right} thora{\sc cost}omy?
\item[A:] (Traumaid prefers) (a left thora{\sc cot}omy).
\end{lexlist}
\stopx
This example suggests that the set that is being considered by the
time the adjective is semantically evaluated is no longer the entire
set including the left and right thoracotomy and thoracostomy
procedures.  In fact, it is not even the set containing only the left
thoracotomy and right thoracostomy procedures, but rather the set
containing only the left thoracotomy procedure, which by definition
does not stand in contrast to any other thoracotomy procedure by
virtue of the property of being performed on the left side.  This set
arises because the noun {\em thoracotomy} restricts over the set including
the left thoracotomy and the right thoracostomy procedures.

To see this, consider the next exchange, uttered in the same
situation.
\startxl{ex:contrast3}
\begin{lexlist}
\item[Q:] Does Traumaid prefer a {\sc left} thora{\sc cot}omy, a {\sc
right} thora{\sc cot}omy or a {\sc left} thora{\sc cost}omy?
\item[A:] (Traumaid prefers) (a {\sc left} thora{\sc cot}omy).
\end{lexlist}
\stopx
Here the set established by the question is restricted by the noun in
the rheme of the answer to be a set of two thoracotomy procedures
(both left and right).  Since they are distinguished by the property
{\em left}, the corresponding linguistic material must be accented.

The algorithm for determining which items are to be stressed for
reasons of contrast works as follows.\footnote{We omit a more detailed
description of the algorithm and its associated data structures for
the sake of brevity.  A more detailed account and numerous examples
are given in Prevost and Steedman (\cite{jcl:93}).} For a given object
$x$, we associate a set of properties which are essential for
constructing an expression that uniquely refers to $x$, as well as a
set of objects (and their referring properties) which might be
considered {\em alternatives} to $x$ with respect to the database
under consideration.  The set of alternatives is restricted by
properties or objects explicitly mentioned in the theme of the
question.  Then for each property of $x$ in turn, we restrict the set
of alternatives to include only those objects having the given
property.  If imposing this restriction decreases the size of the set
of alternatives, then the given property serves to distinguish $x$
{}from its alternatives, suggesting that the corresponding linguistic
material should be stressed.

Besides determining the location of primary sentence stress,
contrastive properties may also necessitate adopting non-standard
lexical stress patterns.  For example, in the following
question/answer pair, the normal lexical stress on {\em thor} switches
to {\em pneu} in {\em pneumothorax} because {\em pneumothorax} stands
in contrast to {\em hemothorax}.
\startxl{ex:lexshift}
\begin{lexlist}
\item[Q:] I know which procedure is recommended for the simple
hemothorax.\\
But which condition is a left {\sc thoracostomy}
recommended for?
\item[A:] A left {\sc thoracostomy} is recommended for the
simple {\sc pneu}mothorax.
\end{lexlist}
\stopx
In the current implementation, such lexical stress shift is handled by
identifying the lexical contrast properties in the alternative set
representations and supplying separate pronunciations in the lexicon.
However, when such properties are determined to stand in contrast to one
another, the alternate pronunciation could in principle be generated by
employing the methods described above within the lexicon.

\section{The Implementation}
\begin{figure}[htb]
\setlength{\unitlength}{0.5cm}
\begin{center}
\begin{picture}(15,22)
\thicklines
\put(4,20){\makebox(7,2){\em{Prosodically Annotated Question}}}
\put(4,17){\framebox(7,2){\sf{Intonational Parser}}}
\put(4,14){\framebox(7,2){\sf{Content Generator}}}
\put(4,11){\framebox(7,2){\sf{CCG Generator}}}
\put(4,8.5){\makebox(7,2){\em{Prosodically Annotated Response}}}
\put(4,6){\framebox(7,2){\sf{Speech Synthesizer}}}
\put(4,3){\makebox(7,2){\em{Spoken Response}}}
\put(13,17){\makebox(7,2){\sf{Discourse Model}}}
\put(16.5,18){\oval(7,2)}
\put(13,14){\makebox(7,2){\sf{Database}}}
\put(16.5,15){\oval(7,2)}
\put(0,14){\makebox(2,2){\sf{CCG}}}
\put(1,15){\oval(4,2)}
\put(7.5,20.5){\vector(0,-1){1.5}}
\multiput(7.5,17)(0,-3){3}{\vector(0,-1){1}}
\multiput(7.5,9)(0,-3){1}{\vector(0,-1){1}}
\put(7.5,6){\vector(0,-1){1.5}}
\thinlines
\put(11,18){\line(1,0){2}}
\put(11,15){\line(1,0){2}}
\put(11,15.5){\line(1,1){2}}
\put(2,14){\line(1,-1){2}}
\put(2,16){\line(1,1){2}}
\end{picture}
\end{center}
\caption{Architecture}
\label{ex:arch}
\end{figure}
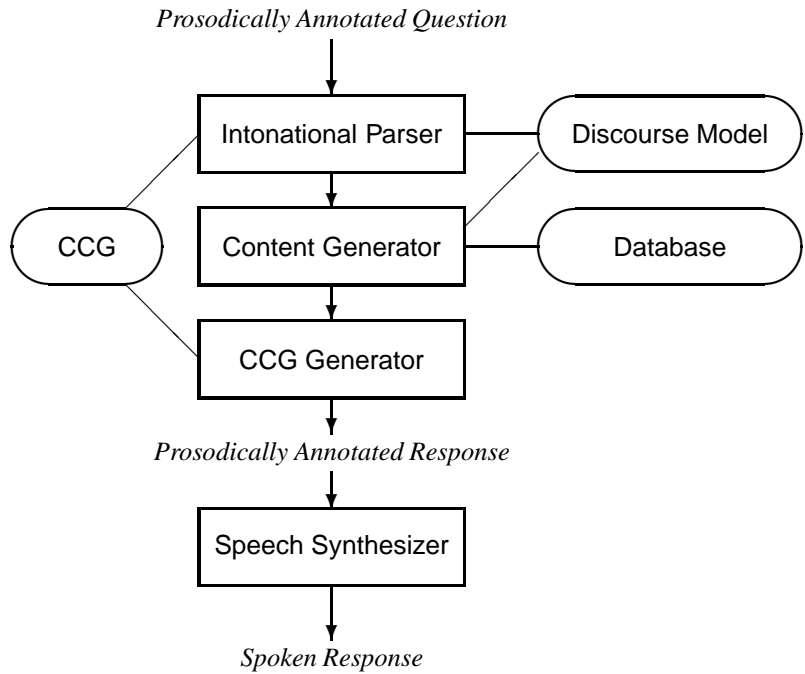

The present paper is an attempt to apply the theories outlined in the
preceding sections to the task of specifying contextually
appropriate intonation for natural language responses to database
queries.  The architecture of the system (shown in Figure~\ref{ex:arch})
identifies the key modules of the system, their relationships to the
database and the underlying grammar, and the dependencies among their
inputs and outputs.

The process begins with a fully segmented and prosodically annotated
representation of a spoken query, as shown in example
\ref{ex:s1}.\footnote{We stress that we do {\em not} start
with a speech wave, but a representation that one might obtain from a
hypothetical system that translates such a wave into strings of words
with Pierrehumbert-style intonation markings.}  We employ a simple
bottom-up shift-reduce parser, making direct use of the combinatory
prosody theory described above, to identify the semantics of the
question.  The inclusion of prosodic categories in the grammar allows
the parser to identify the information structure within the question
as well, marking ``focused'' items with *, as shown in \ref{ex:s2}.
For the moment, unmarked themes are handled by taking the longest
unmarked constituent permitted by the syntax.
\startxl{ex:s1}
\tb{I know what the {\sc cat} scan is for, \\
but {\sc w}\={\sc hich} condi\=tion does {\sc urin}\={\sc alysis} addr\=ess? \\
\> L+H* \> LH\% \> H* \> LL\$}
\stopx
\startxl{ex:s2}
Proposition:\\
$s : \lambda x [condition(x) \& address(*urinalysis, x)]$\\
Theme:\\
$s: \lambda x [condition(x) \& address(*urinalysis, x)]/$\\
\verb#     #$(s:address(*urinalysis, x)/np:x)$\\
Rheme:\\
$s:address(*urinalysis, x)/np:x$
\stopx

The content generation module, which has the task of determining the
semantics and information structure of the response, relies on several
simplifying assumptions.  Foremost among these is the notion that the
rheme of the question is the sole determinant of the theme of the
response, including the specification of focus (although the type of
pitch accent that eventually marks the focus will be different in the
response).  The overall semantic structure of the response can be
determined by instantiating the variable in the lambda expression
corresponding to the {\em wh}-question with a simple Prolog query.
Given the syntactic and focus-marked semantic representation for the
response, along with the syntactic and focus-marked semantic
representation for the theme of the response, a representation for the
rheme of the response can be worked out from the CCG rules.  The
assignment of focus for the rheme of the response (i.e.  the
instantiated variable) must be worked out from scratch, on the basis
of the alternative sets in the database, as described in section 3.

For the question given in \ref{ex:s1}, the content generator
produces the following:
\startxl{ex:s3}
Proposition:\\
$s:address(*urinalysis, *hematuria)$\\
Theme:\\
$s:address(*urinalysis, x)/np:x$\\
Rheme:\\
$np:*hematuria$
\stopx

{}From the output of the content generator, the CCG generation
module produces a string of words and Pierrehumbert-style markings
representing the response, as shown in \ref{ex:s3a}.\footnote{Full
descriptions of the CCG generation algorithm are given in
Prevost and Steedman (\cite{eacl:93,jcl:93}).}
\startxl{ex:s3a}
urinalysis@lhstar\hspace{.5em}addresses@lh\hspace{.5em}hematuria@hstarllb
\stopx
The final aspect of generation involves translating such a string into
a form usable by a suitable speech synthesizer.  The current
implementation uses the Bell Laboratories TTS system (Liberman and
Buchsbaum \cite{att:85}) as a post-processor to synthesize the speech
wave itself.

\section{Results}
The system described above produces quite sharp and natural-sounding
distinctions of intonation contour in minimal pairs of queries like
those in examples \ref{ex:foc1a}--\ref{ex:foc4b}, which should be read
as concerning a single patient with multiple wounds.  These examples
illustrate the system's capability for producing appropriately
different intonation contours for a single string of words under the
control of discourse context. If the responses in these examples are
interchanged, the results sound distinctly unnatural in the given
contexts.\footnote {The first line of each query is for reader
assistance only, and is not processed by the system described here.
The {\em waves} files corresponding to the examples in this section
are available by anonymous ftp from ftp.cis.upenn.edu, under the
directory {\em /pub/prevost/speechcomm}.}

Examples \ref{ex:foc1a} and \ref{ex:foc1b} illustrate the necessity of
the theme/rheme distinction.  Although the pitch accent {\em locations} in
the responses in these examples are identical, occurring on {\em
thoracostomy} and {\em simple}, the alternation in the
theme and rheme tunes is necessary to convey the intended proposition
in the given contexts.

Examples \ref{ex:foc1b} and \ref{ex:foc2b} show that the
system makes appropriate distinctions in focus placement
within themes and rhemes based on context.  Although the responses in these two
sentences possess the same intonational tunes, the pitch accent
location is crucial for conveying the appropriate contrastive
properties.

Examples \ref{ex:foc1a}--\ref{ex:foc4b} manifest the eight basic
combinatorial possibilities for pitch accent placement and tune
selection produced by our program for the given sentence.
The inclusion of contrastive lexical stress shift increases the
number of intonational possibilities even more, as exemplified in
\ref{ex:foc2aa} and \ref{ex:foc2bb}.

\startxll{ex:foc1a}
\begin{lexlist}
\item[Q:] \tb{I know what's recommended for the {\sc
persistent} pneumothorax,\\
but w\=hich proce\=dure \hspace{.4em} is recommended for the {\sc si}\={\sc
mple} pneumoth\=orax?\\
\> L+H* \> LH\% \> H* \> LL\$}
\item[A:] \tb{A left {\sc thorac}\={\sc ost}\={\sc omy} \hspace{.4em} is
recommended for the {\sc si}\={\sc mple} pneumoth\=orax.\\
\> H* \> L \> L+H* \> LH\$}
\end{lexlist}
\stopx
\startxll{ex:foc1b}
\begin{lexlist}
\item[Q:] \tb{I know what's recommended for the {\sc persistent}
pneumothorax,\\
but w\=hich pneumoth\=orax \hspace{.4em} is a left {\sc thorac}\={\sc ostomy}
recommende\=d for?\\
\> L+H* \> LH\% \> H* \> LL\$}
\item[A:] \tb{A left {\sc thorac}\={\sc ostomy} is recommende\=d
for\hspace{.4em} the
{\sc s}\={\sc imple} pneumoth\=orax.\\
\> L+H* \> LH\% \> H* \> LL\$}
\end{lexlist}
\stopx
\startxll{ex:foc2a}
\begin{lexlist}
\item[Q:] \tb{I know what's recommended for the {\sc peritonitis},\\
but w\=hich proce\=dure \hspace{.4em} is recommended for the simple
pneumo{\sc t}\={\sc hor}ax?\\
\> L+H* \> LH\% \> H* LL\$}
\item[A:] \tb{A left {\sc thorac}\={\sc ost}\={\sc omy} \hspace{.4em} is
recommended for the simple pneum\=o{\sc thor}ax.\\
\> H* \> L \> L+H* LH\$}
\end{lexlist}
\stopx
\startxll{ex:foc2b}
\begin{lexlist}
\item[Q:] \tb{I know what's recommended for the {\sc peritonitis},\\
but w\=hich condi\=tion \hspace{.4em} is a left {\sc thorac}\={\sc ostomy}
recommende\=d for?\\
\> L+H* \> LH\% \> H* \> LL\$}
\item[A:] \tb{A left {\sc thorac}\={\sc ostomy} is recommende\=d
for\hspace{.4em} the
simple pneumo\={\sc thor}ax.\\
\> L+H* \> LH\% \> H* LL\$}
\end{lexlist}
\stopx
\startxll{ex:foc3a}
\begin{lexlist}
\item[Q:] \tb{A {\sc right} thoracostomy is recommended for the {\sc
persistent} pneumothorax,\\
but w\=hich thoracost\=omy \hspace{.4em} is recommended for the {\sc s}\={\sc
imple} pneumoth\=orax?\\
\> L+H* \> LH\% \> H* \> LL\$}
\item[A:] \tb{A {\sc l}\={\sc eft} thoracost\=omy \hspace{.4em} is
recommended for the {\sc s}\={\sc imple} pneumoth\=orax.\\
\> H* \> L \> L+H* \> LH\$}
\end{lexlist}
\stopx
\startxll{ex:foc3b}
\begin{lexlist}
\item[Q:] \tb{A {\sc right} thoracostomy is recommended for the {\sc
persistent}
pneumothorax,\\
but w\=hich pneumoth\=orax \hspace{.4em} is a {\sc l}\={\sc eft} thoracostomy
recommende\=d for?\\
\> L+H* \> LH\% \> H* \> LL\$}
\item[A:] \tb{A \={\sc left} thoracostomy is recommende\=d for\hspace{.4em} the
{\sc s}\={\sc imple} pneumoth\=orax.\\
\> L+H* \> LH\% \> H* \> LL\$}
\end{lexlist}
\stopx
\startxll{ex:foc4a}
\begin{lexlist}
\item[Q:] \tb{A {\sc right} thoracostomy is recommended for some
condition,\\
but w\=hich thoracost\=omy \hspace{.4em} is recommended for the simple
pneumo\={\sc thor}ax?\\
\> L+H* \> LH\% \> H* LL\$}
\item[A:] \tb{A {\sc l}\={\sc eft} thoracost\=omy \hspace{.4em} is
recommended for the simple pneum\=o{\sc thor}ax.\\
\> H* \> L \> L+H* LH\$}
\end{lexlist}
\stopx
\startxll{ex:foc4b}
\begin{lexlist}
\item[Q:] \tb{A {\sc right} thoracostomy is recommended for some
condition,\\
but w\=hich condi\=tion \hspace{.4em} is a {\sc l}\={\sc eft}
thoracostomy recommende\=d for?\\
\> L+H* \> LH\% \> H* \> LL\$}
\item[A:] \tb{A \={\sc left} thoracostomy is recommende\=d for\hspace{.4em} the
simple pneumo\={\sc thor}ax.\\
\> L+H* \> LH\% \> H* LL\$}
\end{lexlist}
\stopx
\startxll{ex:foc2aa}
\begin{lexlist}
\item[Q:] \tb{I know which procedure is recommended for the simple
hemothorax,\\
but w\=hich proce\=dure \hspace{.4em} is recommended for the simple
{\sc p}\={\sc neu}moth\=orax?\\
\> L+H* \> LH\% \> H* \> LL\$}
\item[A:] \tb{A left {\sc thorac}\={\sc ost}\={\sc omy} \hspace{.4em} is
recommended for the simple \={\sc pneu}moth\=orax.\\
\> H* \> L \> L+H* \> LH\$}
\end{lexlist}
\stopx
\startxll{ex:foc2bb}
\begin{lexlist}
\item[Q:] \tb{I know which procedure is recommended for the simple
hemothorax,\\
but w\=hich condi\=tion \hspace{.4em} is a left {\sc thorac}\={\sc ostomy}
recommende\=d for?\\
\> L+H* \> LH\% \> H* \> LL\$}
\item[A:] \tb{A left {\sc thorac}\={\sc ostomy} is recommende\=d
for\hspace{.4em} the
simple {\sc p}\={\sc neu}moth\=orax.\\
\> L+H* \> LH\% \> H* \> LL\$}
\end{lexlist}
\stopx

\section{Conclusions}
The results show that is possible to generate synthesized spoken
responses with contextually appropriate intonational contours in a
database query task.  Many important problems remain, both because of
the limited range of discourse-types and intonational tunes considered
here, and because of the extreme oversimplification of the discourse
model (particularly with respect to the ontology, or variety of types
of discourse entities).  Nevertheless, the system presented here has a
number of properties that we believe augur well for its extension to
richer varieties of discourse, including the types of monologues and
commentaries that are more appropriate for the actual TraumAID domain.
Foremost among these is the fact that the system and the underlying
theory are entirely modular.  That is, any of its components can be
replaced without affecting any other component because each is
entirely independent of the particular grammar defined by the lexicon
and the particular knowledge base that the discourse concerns.  It is
only because CCG allows us to unify the structures implicated
in syntax and semantics on the one hand, and intonation and discourse
information on the other, that this modular structure can be so simply
attained.

\baresection{Acknowledgments}
Preliminary versions of some sections in the present paper were
published as Prevost and Steedman (\cite{eacl:93,euro:93}).  We are
grateful to the audiences at those meetings, to AT\&T Bell
Laboratories for allowing us access to the TTS speech synthesizer, to
Mark Beutnagel, Julia Hirschberg, and Richard Sproat for patient
advice on its use, to Abigail Gertner for advice on Traumaid, to
Janet Pierrehumbert for discussions on notation, and to the anonymous
referees for many helpful suggestions.  The usual
disclaimers apply.  The research was supported in part by NSF grant
nos. IRI90-18513, IRI90-16592, IRI91-17110 and CISE IIP-CDA-88-22719,
DARPA grant no.  N00014-90-J-1863, ARO grant no.  DAAL03-89-C0031, and
grant no.  R01-LM05217 from the National Library of Medicine.


\newpage
\baresection{Bibliography}
\ \vspace{-.2in}
\begin{description}
\addtolength{\itemsep}{-8pt}
\small

\bibitem{1986}{BandP:86}
M. Beckman and J. Pierrehumbert (1986), ``Intonational Structure
in Japanese and English'', {\em Phonology Yearbook}, Vol. 3, pp. 255--310.

\bibitem{1991}{Bird:91}
S. Bird (1991), ``Focus and phrasing in Unification Categorial
Grammar'', {\em Declarative Perspectives on Phonology}, Working Papers
in Cognitive Science 7, ed. by S. Bird (University of Edinburgh), pp.
139--166.

\bibitem{1988}{Davi:88}
J. Davis and J. Hirschberg (1988), ``Assigning Intonational
Features in Synthesized Spoken Directions'', {\em Proceedings of
the 26th Annual Conference of the ACL}, Buffalo, pp. 187--193.

\bibitem{1990}{Hirs:90}
J. Hirschberg (1990), ``Accent and Discourse Context:  Assigning
Pitch Accent in Synthetic Speech'', {\em Proceedings of AAAI: 1990},
pp. 952--957.

\bibitem{1986}{Houg:86}
G. Houghton (1986), {\em The Production of Language in Dialogue: a
Computational Model}, unpublished PhD dissertation, University of Sussex.

\bibitem{1988}{Isar:88}
S. Isard and M. Pearson (1988), ``A Repertoire of British English
Intonation Contours for  Synthetic Speech'', {\em Proceedings of Speech
'88, 7th FASE Symposium}, Edinburgh, pp. 1233--1240.

\bibitem{1985}{att:85}
M. Liberman and A.L. Buchsbaum (1985), ``Structure and Usage of
Current Bell Labs Text to Speech Programs'', TM 11225-850731-11, AT\&T
Bell Laboratories.

\bibitem{1989}{Moor:89}
M. Moortgat (1989), {\em Categorial Investigations} (Foris, Dordrecht).

\bibitem{1988}{Oehr:88}
R. Oehrle (1988), ``Multi-dimensional Compositional Functions as a basis
for Grammatical Analysis'', in {\em Categorial Grammars and
Natural Language Structures}, ed. by R. Oehrle, E. Bach and D.
Wheeler (Reidel, Dordrecht), pp. 349--390.

\bibitem{1980}{Pier:80}
J. Pierrehumbert (1980), {\em The Phonology and Phonetics of English
Intonation}, PhD dissertation, MIT. (Dist. by
Indiana University Linguistics Club,
Bloomington, IN.)

\bibitem{1990}{PandH:90}
J. Pierrehumbert and J. Hirschberg (1990), ``The Meaning of
Intonational Contours in the Interpretation of Discourse'', in {\em
Intentions in Communication}, ed. by P.  Cohen, J. Morgan, and M.
Pollack (MIT Press, Cambridge MA), pp. 271--312.

\bibitem{1993}{prop:93}
S. Prevost (1993), ``Intonation, Context and Contrastiveness in Spoken
Language Generation'', dissertation proposal, University of Pennsylvania.

\bibitem{1993a}{eacl:93}
S. Prevost and M. Steedman (1993a), ``Generating Contextually
Appropriate Intonation'', {\em Proceedings of the Sixth Conference
of the European Chapter of the Association for Computational
Linguistics}, Utrecht, pp. 332--340.

\bibitem{1993b}{euro:93}
S. Prevost, and M. Steedman (1993b), ``Using Context to Specify
Intonation in Speech Synthesis'', {\em Proceedings of the 3rd European
Conference of Speech Communication and Technology (EUROSPEECH)},
Berlin, September 1993, pp. 2103--2106.

\bibitem{1993c}{jcl:93}
S. Prevost and M. Steedman (1993c), ``Generating Intonation from
Context Using a Combinatory Grammar'', manuscript,
University of Pennsylvania.

\bibitem{1987}{cgpg}
M. Steedman (1987). ``Combinatory Grammars and Parasitic Gaps'', {\em
Natural Language and Linguistic Theory}, Vol. 5, pp. 403--439.

\bibitem{1990a}{gacc}
M. Steedman (1990a). ``Gapping as Constituent Coordination'',
{\em Linguistics \& Philosophy}, Vol. 13, pp. 207--263.

\bibitem{1990b}{acl90}
M. Steedman (1990b), ``Structure and Intonation in Spoken Language
Understanding'',
{\em Proceedings of the 25th Annual Conference of the Association for
Computational Linguistics},
Pittsburgh, June 1990, pp. 9--17.

\bibitem{1991a}{sandi}
M. Steedman (1991a), ``Structure and Intonation'', {\em Language}, Vol.
68, pp. 260--296.

\bibitem{1991b}{trandd}
M. Steedman (1991b), ``Type-raising and Directionality in Categorial Grammar'',
{\em Proceedings of the 29th Annual Meeting of the Association for
Computational Linguistics}, Berkeley, June 1991, pp. 71--78.

\bibitem{1991c}{brussels}
M. Steedman (1991c), ``Surface Structure, Intonation, and `Focus' '',
in {\em Natural Language and Speech, Proceedings of the ESPRIT
Symposium, Brussels, 1991}, ed. by E. Klein and F. Veltman, pp. 21--38.

\bibitem{1984}{Terk:84}
J. Terken (1984), ``The Distribution of Accents in
Instructions as a Function of Discourse Structure'', {\em Language and
Speech}, Vol 27, pp. 269--289.

\bibitem{1992}{webb:92}
B. Webber, R. Rymon and J.R. Clarke (1992), ``Flexible Support for Trauma
Management through Goal-directed Reasoning and Planning'',
{\em Artificial Intelligence in Medicine}, Vol. 4(2), pp. 145-163.

\bibitem{1979}{Youn:79}
S. Young and F. Fallside (1979), ``Speech Synthesis from Concept: a
Method for Speech Output from Information Systems'', {\em Journal of the
Acoustical Society of America}, Vol. 66, pp. 685--695.

\bibitem{1993}{Zach:93}
R. Zacharski, A.I.C. Monaghan, D.R. Ladd and J. Delin (1993), ``BRIDGE:
Basic Research on Intonation in Dialogue Generation'', unpublished
manuscript. HCRC, University of Edinburgh.

\end{description}

\end{document}